\documentclass{article}

\usepackage{PRIMEarxiv}

\usepackage[utf8]{inputenc} 
\usepackage[T1]{fontenc}    
\usepackage{hyperref}       
\usepackage{url}            
\usepackage{booktabs}       
\usepackage{amsfonts}       
\usepackage{nicefrac}       
\usepackage{microtype}      
\usepackage{lipsum}
\usepackage{graphicx}
\graphicspath{{media/}}     

\usepackage{txfonts}
\usepackage{graphicx}
\usepackage{epstopdf}
\usepackage{multirow}
\usepackage{booktabs}

\usepackage{algorithm}
\usepackage{algorithmic}

\title{A Safe Semi-supervised Graph Convolution Network}

\author{
  Zhi Yang \\
  School of Computer Science \\
  Hubei University of Technology \\
  Wuhan China\\
  \texttt{zyang631@hbut.edu.cn} \\
  \And
  Yadong Yan \\
  School of Computer Science \\
  Hubei University of Technology \\
  Wuhan China\\
  \texttt{913176990@qq.coml} \\
  \AND
  Haitao Gan \\
  School of Computer Science \\
  Hubei University of Technology \\
  Wuhan China\\
  \texttt{email} \\
  \And
  Jing Zhao \\
  School of Life Sciences\\
  Hubei University\\
  Wuhan China \\
  \texttt{zhaojing@hubu.edu.cn} \\
  \And
  Zhiwei Ye \\
  School of Computer Science \\
  Hubei University of Technology \\
  Wuhan China\\
  \texttt{27454010@qq.com} \\
  }

\begin{document}

\maketitle

\begin{abstract}
In the semi-supervised learning field, Graph Convolution Network (GCN), as a variant model of GNN, has achieved promising results for non-Euclidean data by introducing convolution into GNN. However, GCN and its variant models fail to safely use the information of risk unlabeled data, which will degrade the performance of semi-supervised learning. Therefore, we propose a Safe GCN framework (Safe-GCN) to improve the learning performance. In the Safe-GCN, we design an iterative process to label the unlabeled data. In each iteration, a GCN and its supervised version(S-GCN) are learned to find the unlabeled data with high confidence. The high-confidence unlabeled data and their pseudo labels are then added to the label set. Finally, both added unlabeled data and labeled ones are used to train a S-GCN which can achieve the safe exploration of the risk unlabeled data and enable safe use of large numbers of unlabeled data. The performance of Safe-GCN is evaluated on three well-known citation network datasets and the obtained results demonstrate the effectiveness of the proposed framework over several graph-based semi-supervised learning methods.
\end{abstract}


\section{Introduction}
In recent years, graph-based methods have attracted more and more attention from researchers. The reason is twofold. First, in the real-world problem, there exists a large amount of non-Euclidean data, such as recommender systems \cite{luo2020motif}, proteins-proteins network\cite{fout2017protein,wang2020deep,zhang2021graph}, etc. Unlike the Euclidean data, non-Euclidean data have an irregular data structure, these data can be expressed by graph for the powerful ability of graph. Second, the geometric structure of data can be embedded by graph analysis methods, thereby helping the model to improve its recognition ability. In general, graph-based methods extend the application scenarios of existing machine learning methods to a certain extent.\

Nowadays, graph neural networks (GNNs) have been widely used in machine learning for their convincing performance and high interpretability\cite{zhou2020graph}. Due to the great success of CNNs in machine learning, a large number of researchers have tried to extend convolution operations from the image domain to the graph domain\cite{bruna2013spectral, henaff2015deep,atwood2016diffusion,kipf2016semi}. As for graph-based semi-supervised learning, \cite{kipf2016semi}proposed graph convolution network (GCN). GCN used convolution on the graph to extract features and obtained both feature information and graph structure information of the nodes. However, it required the full graph Laplacian operator when training the model, which was very computationally intensive when computing large graphs. Moreover, the embedding of nodes in each layer of GCN was recursively computed by the embedding of all neighboring nodes in the previous layer. This made the receptive field of the nodes grow exponentially with the number of layers, which could be very time-consuming when computing the gradient. GraphSAGE\cite{hamilton2017inductive} replaced the graph Laplacian function of GCN with a learnable aggregation function and used neighbor sampling to reduce the growth of the receptive domain, making it applicable to inductive learning. FastGCN \cite{chen2018fastgcn} interpreted the graph convolution as an integral transform of the embedding function under probability measures and used important sampling to reduce the computational time while providing a comparable level of computational accuracy. SGC \cite{wu2019simplifying} reduced the additional complexity by successively removing the nonlinearity in the GCN layers and collapsing the resulting function into a linear transformation. The same computational speedup could be achieved without negatively affecting the classification accuracy. Different from GCN, TAGCN \cite{du2017topology} used K graph convolution kernels at each layer to extract local features of different sizes, following the example of CNNs, to avoid the previous drawback of approximating the convolution kernels without extracting complete and sufficient graph information, and to improve the representational power of the model. GAT \cite{velivckovic2017graph} made use of hidden self-attentive layers by stacking the nodes in each layer so that it can focus on the features of the neighborhood and assign different weights to different nodes in the neighborhood, making it possible to achieve advanced results without the need to know the graph structure in advance. It addressed the problem that GCN needs to have prior knowledge of unknown data and the usage scenarios are mostly in dealing with static graphs. SuperGAT\cite{kim2020find} learned more appropriate attention weights when distinguishing misconnected neighbors by a self-supervised task. Many different improved versions have been proposed by researchers since then\cite{chiang2019cluster,pei2020geom,yu2020forecasting}, and all have achieved promising results. \

On the other hand, GCN and its variant models are mostly built on a semi-supervised learning paradigm\cite{chapelle2009semi}. Semi-supervised learning used labeled data to train the model while using unlabeled data to better maintain the intrinsic structural information of the data, allowing the model to achieve promising results. And it is well adapted to problems that contain a small number of labeled data and a large number of unlabeled data. Nevertheless, it is known that there are risky unlabeled data in unlabeled dataset, which may degrade the performance of the model. In some cases, semi-supervised learning performs worse than the corresponding supervised learning, as has been verified in many works \cite{singh2008unlabeled,chawla2005learning, gan2013semi}. If the risk of unlabeled data cannot be reduced, this will make the model very limited for using in practical scenarios. Therefore, it is necessary to design a safe GCN.\

Some recent works have used the self-training method\cite{li2018deeper,sun2020multi,zhou2019dynamic,pedronette2021rank}  to select the unlabeled data and expanded the labeled dataset by the high-confidence unlabeled data, which somewhat alleviates the risk problem of unlabeled data. The main idea of Self-training is using the labeled data to train a classifier to label the unlabeled data\cite{chapelle2009semi,scudder1965probability}, then select some high-confidence data to expand the labeled dataset. This process is carried out iteratively until convergence. How to design the Self-training method to select reasonable unlabeled data is an important challenge. Most works are only based on the highest soft-max output\cite{sun2020multi,zhou2019dynamic}, which is always insufficient to measure the confidence of the data.\

In this paper, we propose a safe GCN framework (Safe-GCN). The proposed model is implemented in three stages. First, S-GCN and GCN classifiers are trained to obtain the pseudo-label of unlabeled data. Second, the outputs of S-GCN and GCN are compared. The unlabeled data with high- confidence are selected by a confidence filtering condition. Then labeled dataset is expanded in a balanced way by high-confidence unlabeled data. Finally, the expanded labeled dataset is used to train the S-GCN. Hence, our proposed Safe-GCN makes better use of supervised and semi-supervised information, and has the opportunity to achieve model security by reducing the negative impact of risky unlabeled data.\

We conducted experiments on three publicly available citation datasets. The results demonstrate that the classification performance of the proposed model outperforms most existing graph-based models. The main contributions and advantages of this paper compared to related works are summarized as follows:
\begin{itemize}
\item  We proposed a safe semi-supervised graph convolution model that can effectively reduce the adverse effects of risky unlabeled samples. The model can safely utilize a large amount of unlabeled data.\
\item	 The proposed model utilizes only the information of the training data during training and does not need to know the graph structure and feature information of the test data. Therefore, the proposed model can be directly applicable to inductive learning.

\end{itemize}

The rest of the paper is organized as follows. Section 2 describes the background in this paper. Section 3 describes the proposed algorithm. The dataset, experimental configuration and results are described in Section 4. Section 5 gives the conclusion of this paper and discusses future directions.

\section{Background}

Since GCN and its supervised version are used as classifiers in Safe-GCN, so we discuss the details of GCN in this section.\
(Kipf \& Welling, 2017)\cite{kipf2016semi}proposed a graph convolution network for semi-supervised learning. The main idea is to pass information from each node to its neighbors through information transfer between nodes, and iteratively aggregate the features of the nodes' neighbors through Laplace matrix and convolution on the graph, enabling it to deeply estimate the labels of unlabeled data. The model can be described as following:
$$F=f(X,\ A)$$

where F$\in\mathbb{R}^{n\times d}$ denotes a label matrix that represents the output of the unlabeled data. X is the feature matrix of the dataset and A is the adjacency matrix associated with data. \

The propagation law of the layers in GCN is given by:
$$H^{i+1}=\ \sigma(D^{-\frac{1}{2}}\widetilde{A}D^{-\frac{1}{2}}H^{\left(i\right)}W^{(i)})$$
where $\widetilde{A}=A+I$ denotes the A matrix with added self-connections, I denotes the identity matrix, and $\widetilde{D}$ denotes the degree matrix of the $\widetilde{A}$ matrix. $W^{(i)}$ denotes the weight matrix corresponding to the ith layer of the network. $\sigma(\cdot)$ denotes the activation function, it is given by the ReLU. $H^{(0)}$ is X. \

Since the GCN model can achieve advanced results with 2-3 layers, a two-layer GCN model has the following form:
$$F=softmax(\breve{A}\sigma\left(\breve{A}W^{(0)}X\right)W^{(1)})$$
where $\breve{A}=D^{-\frac{1}{2}}\widetilde{A}D^{-\frac{1}{2}}$ denotes the regularized Laplace matrix, $W^{(0)}$ is the input-hidden weight matrix, and $W^{(1)}$ is the hidden-output weight matrix. \

The softmax activation function converts the output matrix into a probability distribution for each data corresponding to each category by row, i.e., the probability of each data corresponding to all categories sums to 1. Deep neural networks is learned by making the predicted label as close as possible to the ground-truth label. This is achieved by minimizing the cross-entropy loss function that is typically used for classification problems. The cross-entropy function is used as the loss function in the GCN.

\section{Our algorithm}
As stated in Section 1, the GCN and its variants models use unlabeled data to enhance the performance of the model and achieve promising results. However, mistakes in risky unlabeled data can spread during the training of the model and may degrade the performance of the model, which makes the use of unlabeled data very risky. Considering the safe use of unlabeled data, we propose an enhanced safe GCN model, which is based on the self-training framework. The method to select high-confidence data in the proposed model is illustrated in Fig \ref{fig:1}.  The model is divided into three stages as shown below:\

\begin{itemize}
\item[(1)] Pseudo label acquisition: the first stage computes the information encoded in the embedding by S-GCN and GCN to make better use of labeled and unlabeled data. The classification results and model outputs of S-GCN and GCN for unlabeled data are obtained in this stage.

\item[(2)] Labeled dataset expansion: the second stage evaluates the unlabeled data and performed data expansion. Most self-training methods use high-confidence unlabeled data to expand the labeled dataset, and they rely on the maximum softmax scores to assess the risk level of unlabeled data \cite{zhou2019dynamic,pedronette2021rank},which is always not accurate enough. Therefore, we propose a new confidence-based data filtering condition. Meanwhile, the same number of unlabeled data with high confidence for each class is added to the labeled dataset to ensure a balanced distribution of labels. The first and second stages of learning are iteratively performed until the stopping condition is satisfied(i.e. there are no data meeting filtering condition in the unlabeled dataset).

\item[(3)]S-GCN classification: Supervised GCN learning is performed using the final expanded dataset to predict the test data and obtain the final results.

\end{itemize}

\begin{figure}[ht]
\centering
\includegraphics[width=\linewidth]{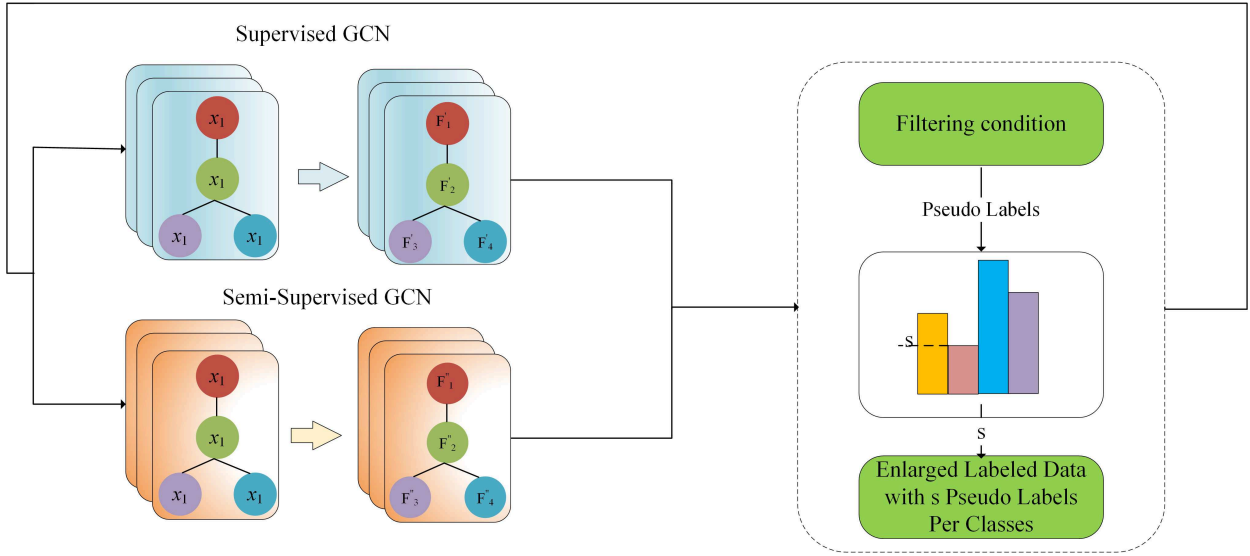}
\caption{Flowchart of high-confidence data expansion in Safe-GCN}
\label{fig:1}
\end{figure}

\subsection*{Pseudo-label acquisition}

This stage is the fundamental module that obtained pseudo label of unlabeled data by S-GCN and GCN. \

The feature matrix of the dataset is defined as $X=\left[x_1,\ldots,x_l,\ldots,x_n\right], X\in\mathbb{R}^{n\times d}$\ , n is the number of data, d is the number of feature dimensions of the data. l denotes the total number of initially labeled data. The number of labeled data increases with the iterations, and the details are given in the following sections. The feature matrix of the initial labeled data is defined as: $X^{l\left(0\right)}=\left[x_1,\ldots,x_l\right]$ with the corresponding initial label set $Y^{l\left(0\right)}=\left[y_1,...\ ,\ y_l\right]^T$, each labeled data $x_p$ will have a label $y_p\ \in\ {1,\ \ldots\ ,\ c}$ with the number of classes c. And the labeled dataset after the kth iteration is $X^{l\left(k\right)}$,

The network structure of S-GCN is the same as the GCN, with the difference: 1) the adjacency matrix and degree matrix are constructed using labeled data. 2) only labeled data are involved in model training. The model output of the kth iteration is expressed as follow:
$$F^{\prime\left(k\right)}={\ f}_{S-GCN}(X^{l\left(k\right)},\ A^{\prime(k)},Y^{l(k)})$$

$A^{\prime\left(k\right)}\in\mathbb{R}^{l\times l}$ denotes the adjacency matrix of the S-GCN after the kth iteration. The model output of the kth iteration of the GCN is given by:
$$F^{\prime\prime\left(k\right)}=f_{GCN}(X,A^{\prime\prime},Y^{l(k)})$$

where $A^{\prime\prime}\in\mathbb{R}^{(l+u)\times(l+u)}$ denotes the adjacency matrix of the GCN.  \

Formally the training dataset can be represented by a labeling function with the goal of learning the labeling functions $L(F^{\prime\left(k\right)}), L(F^{\prime\prime\left(k\right)})$ in S-GCN and GCN.

\subsection*{Labeled dataset expansion}

Once the data have been processed by the softmax layer, we are interested in the largest value in $\mathbf{F}_\mathbf{i}^{\prime\left(k\right)}$,$\mathbf{F}_\mathbf{i}^{\prime\prime\left(k\right)}$ that is associated with the most likely class. Therefore, we consider a function $m(\cdot)$ that returns the largest element in the vector. The function can be defined as: $m(\cdot)=max(\cdot)$, $m(\mathbf{F}_\mathbf{i}^{\prime\left(k\right)})$ returns the largest element in $\mathbf{F}_\mathbf{i}^{\prime\left(k\right)}$. $\mathbf{F}_\mathbf{i}^{\prime\left(k\right)}$, $\mathbf{F}_\mathbf{i}^{\prime\prime\left(k\right)}$ denotes the output vector of the ith data in S-GCN and GCN, respectively, both being c-dimensional vector. \

However, the maximum value of the output vector of the softmax layer is not sufficient to determine the security of data. To this end, we propose a confidence-based filtering condition for unlabeled data:(1) the prediction results of S-GCN and GCN for the unlabeled data are the same(2) the maximum value of the output vector of GCN is greater than or equal to the S-GCN and greater than the confidence threshold $\alpha$. We treat data that satisfy the filtering condition as high confidence ones. \

Labels of data that meet the conditions form a candidate label set. The histogram is used to count the alternative label set. In order to balance the distribution of label set, the number of data per class is counted in a histogram. Formally, a function  $hgram(\cdot)$ is defined to statistically the classes in the label set and the number of labels in each class. $s^{(k)}$ denotes the number of labels of the least labeled classes in the candidate label set for the kth iteration. If there is a class that is not present in the candidate label set, the class label is not updated

\subsection*{Supervised GCN classification}

The first and second stages are iteratively applied until there are no more unlabeled data satisfying the filtering condition. The labeled dataset after the completion of the iteration is the final labeled dataset. The third stage uses S-GCN model to learn the expanded dataset and obtain the final classification result, i.e., the classification result of Safe-GCN.

The three stages of the proposed method are detailed and formally defined in the following subsections. An overview of the proposed method is given in Algorithm \ref{algo1}. Line 1 and 2 of the algorithm table train a GCN and an S-GCN, respectively, and lines 3 to 9 select high-confidence data by filtering conditions. Lines 10 to 15 add the high-confidence data to the labeled dataset in a balanced way.

\begin{algorithm}[ht]
\caption{Safe-GCN}\label{algo1}
\begin{algorithmic}[1]
\REQUIRE Feature matrix X, labeled data adjacency matrix $A^{\prime(0)}$, train data adjacency matrix $A^{\prime\prime}$, initial labeled data $X^{l(0)}$ with the corresponding labels $Y^{l(0)}$, initial unlabeled data $X^{u\left(0\right)}$, confidence threshold $\alpha$
\ENSURE S-GCN Embedding matrix $\mathbf{F}^{\prime\left(k\right)}$
\FOR {each stage k}
      \STATE $\mathbf{F}^{\prime\left(k-1\right)}={\ f}_{S-GCN}(X^{l\left(k-1\right)},\ A^{\prime(k-1)},Y^{l(k-1)})$
      \STATE $\mathbf{F}^{\prime\prime(k-1)}=f_{GCN}(X,A^{\prime\prime},\ Y^{l(k-1)})$
      \FOR {all $x_i\in\ X^{u\left(k-1\right)}$}
          \IF {$m(\mathbf{F}_\mathbf{i}^{\prime\prime(\mathbf{k}-\mathbf{1})})\geq m(\mathbf{F}_\mathbf{i}^{\prime(\mathbf{k}-\mathbf{1})})\geq \alpha$
          $\mathbf{and}\ L(\mathbf{F}_\mathbf{i}^{\prime(\mathbf{k}-\mathbf{1})})=\ L(\mathbf{F}_\mathbf{i}^{\prime\prime(\mathbf{k}-\mathbf{1})})$}
          \STATE $y_i=L(\mathbf{F}_\mathbf{i}^{\prime\prime(\mathbf{k}-\mathbf{1})})$
          \STATE ${\ddot{Y}}={\ddot{Y}}\cup y_i$, ${\ddot{Y}}$ is the candidate label set.
          \ENDIF
      \ENDFOR
      \STATE $s^{(k-1)}=min(hgram(\ \ddot{Y}))$
      \FOR {each class in ${\ddot{Y}}$}
           \STATE Update $Y^{l(k-1)}$ with the top $s^{(k-1)}$ labels.
           \STATE Add the corresponding data to $X^{l\left(k-1\right)}$.
           \STATE Delete the corresponding data from the Unlabeled dataset $X^{u\left(k-1\right)}$.
      \ENDFOR
      \STATE Clear $\ddot{Y}$
\ENDFOR \\
\RETURN { $\mathbf{F}^{\prime\left(k\right)}$}
\end{algorithmic}
\end{algorithm}

\section{Experimental evaluation}
\subsection*{Citation dataset}

The predictive power of the model is evaluated on three citation network datasets: Cora, Citeseer and Pubmed \cite{sen2008collective}. These datasets have been utilized in many graph-based semi-supervised classification tasks. The division of datasets are shown in Table \ref{tab:1}, and a brief introduction of datasets are as follows:

\paragraph{Cora} The Cora dataset consists of 2708 scientific publications, each publication is described by a 1433-dimensional word vector with values of 0 and 1, respectively, representing whether corresponding word appears in the paper. Publications of Cora are divided into 7 classes. The division of Cora is following the GCN. The difference is that we use the union of the train set and the validation set as our train set.

\paragraph{Citeseer} The Citeseer dataset employs a similar representation to Cora, but the publications are divided into 6 classes and the data are described by a 3703-dimensional word vector. The division of Citeseer is also following the GCN.

\paragraph{Pubmed} The dataset includes 19717 scientific publications on diabetes from the Pubmed database. Publications are divided into three classes and described by a TF/IDF-weighted word vector in a dictionary of 500 unique words. \

In many practical applications of machine learning, the information of the test data during the training of the model is unknown. Therefore, the three citation datasets are divided differently from those in GCN, the model is trained without using feature information and node information from the test dataset. For a fair comparison, the methods used for comparison also use the same form of data division (i.e., feature and node information from the test dataset is not used). The initial labeled data is trained using 20 labels per class. The specific division is shown in Table \ref{tab:1}.

\begin{table}[ht]
\begin{center}
\caption{\label{tab:1}Citation network datasets statistics}
\setlength{\tabcolsep}{7mm}{
\begin{tabular}{@{}lllll@{}}
\toprule
Dataset  & Nodes & Labels & Train & Test\\
\midrule
Cora     & 2708  & 140    & 1708  & 1000  \\
Citeseer & 3327  & 120    & 2327  & 1000  \\
Pubmed   & 19717 & 60     & 18170 & 1000  \\
\bottomrule
\end{tabular}}
\end{center}
\end{table}

\subsection*{Experimental setup}

We compare the proposed model Safe-GCN with some traditional machine learning methods and some state-of-the-art graph-based methods. These methods belong to two categories: (1) traditional machine learning algorithms. (2) graph-based convolution networks. \

The traditional machine learning algorithms include multilayer perceptron (MLP) and support vector machine (SVM). The graph-based models include representative semi-supervised graph convolution networks (GCN) \cite{kipf2016semi}, Graph Attention Network (GAT) \cite{velivckovic2017graph}, Topology Adaptive Convolutional Network (TAGCN) \cite{du2017topology}, Predict then Propagate: Graph Neural Networks meet Personalized PageRank \cite{klicpera2018predict} and Attention-based Graph Neural Network for Semi-supervised Learning \cite{thekumparampil2018attention}.\

The implementation of the proposed Safe-GCN and the above methods were made upon the Pytorch framework. The graph-based methods were implemented via Pytorch Geometric (PYG), an extension library for geometric learning based on the Pytorch framework, and the traditional machine learning methods were implemented via the Scikit-learn package.\

Adam was used to training all the above graph-based models as optimizers. During the training phase, each model's hyperparameters and network configuration were followed the default benchmark provided in PYG. The learning rate of the models was set to 0.01 and the dropout parameter was defined as 0.5, except for GAT which was 0.6. For a fair comparison, the number of epochs of all models was limited to 200. MLP and SVM followed the default setups in Scikit-learn, the maximum number of iterations were defined as 1000 for MLP to ensure full convergence.

\subsection*{Method comparison}
Table \ref{tab:2} illustrates the classification rate using different traditional machine learning and graph-based methods for Cora, Citeseer, and Pubmed datasets(following the split in \cite{kipf2016semi}). Table \ref{tab:3} illustrates the average classification rate(together with its standard deviation over the ten random splits). For each table, there are three columns that correspond to three citation datasets.

\begin{table}[ht]
\begin{center}
\caption{\label{tab:2}Results of multiple citation network dataset}
\setlength{\tabcolsep}{12mm}{
\begin{tabular}{@{}llll@{}}
\toprule
Method              & Cora           & Citeseer        & Pubmed          \\
\midrule
SVM           & 0.5550          & 0.5842          & 0.7143          \\
MLP           & 0.5270          & 0.4985          & 0.6988          \\
S-GCN         & 0.6120          & 0.5941          & 0.6833          \\
GCN           & 0.7170          & 0.6512          & 0.7479          \\
TAGCN         & 0.6400          & 0.5409          & 0.5908          \\
GAT           & 0.7370          & 0.6522          & 0.7589          \\
APPNP         & 0.7560          & 0.6600          & 0.7430          \\
AGNN          & 0.7540          & 0.6610          & 0.7770          \\
\textbf{Safe-GCN} & \textbf{0.7630}  & \textbf{0.6985} & \textbf{0.7776} \\
\bottomrule
\end{tabular}}
\end{center}
\end{table}

\begin{table}[ht]
\begin{center}
\caption{\label{tab:3}Recognition performance (Mean recognition accuracy $ \pm $ Standard deviation on multiple citation network dataset over 10 different random splits.}
\setlength{\tabcolsep}{12mm}{
\begin{tabular}{@{}llll@{}}
\toprule
Method              & Cora           & Citeseer        & Pubmed          \\
\midrule
SVM           & $0.5483 \pm 0.018$         & $0.5495 \pm 0.032$         & $0.6874 \pm 0.023$          \\
MLP           & $0.4992 \pm 0.021$         & $0.4581 \pm 0.054$         & $0.6722 \pm 0.019$         \\
S-GCN         & $0.6507 \pm 0.019$         & $0.5989 \pm 0.031$         & $0.7039 \pm 0.022$          \\
GCN           & $0.7009 \pm 0.017$         & $0.6449 \pm 0.022$         & $0.7424 \pm 0.018$          \\
TAGCN         & $0.6271 \pm 0.029$         & $0.4987 \pm 0.035$         & $0.6120 \pm 0.060$          \\
GAT           & $0.7160 \pm 0.017$         & $0.6506 \pm 0.016$         & $0.7425 \pm 0.023$          \\
APPNP         & $0.7204 \pm 0.026$         & $0.6562 \pm 0.018$         & $0.7570 \pm 0.022$         \\
AGNN          & $0.7161 \pm 0.025$         & $0.6558 \pm 0.010$         & $0.7564 \pm 0.017$          \\
\textbf{Safe-GCN} & $\textbf{0.7345} \pm \textbf{0.015}$  & $\textbf{0.6845} \pm \textbf{0.017}$ & $\textbf{0.7799} \pm \textbf{0.020}$ \\
\bottomrule
\end{tabular}}
\end{center}
\end{table}

\subsection*{The effect of the number of labeled data}

\begin{figure}[ht]
\centering
\includegraphics[width=\linewidth]{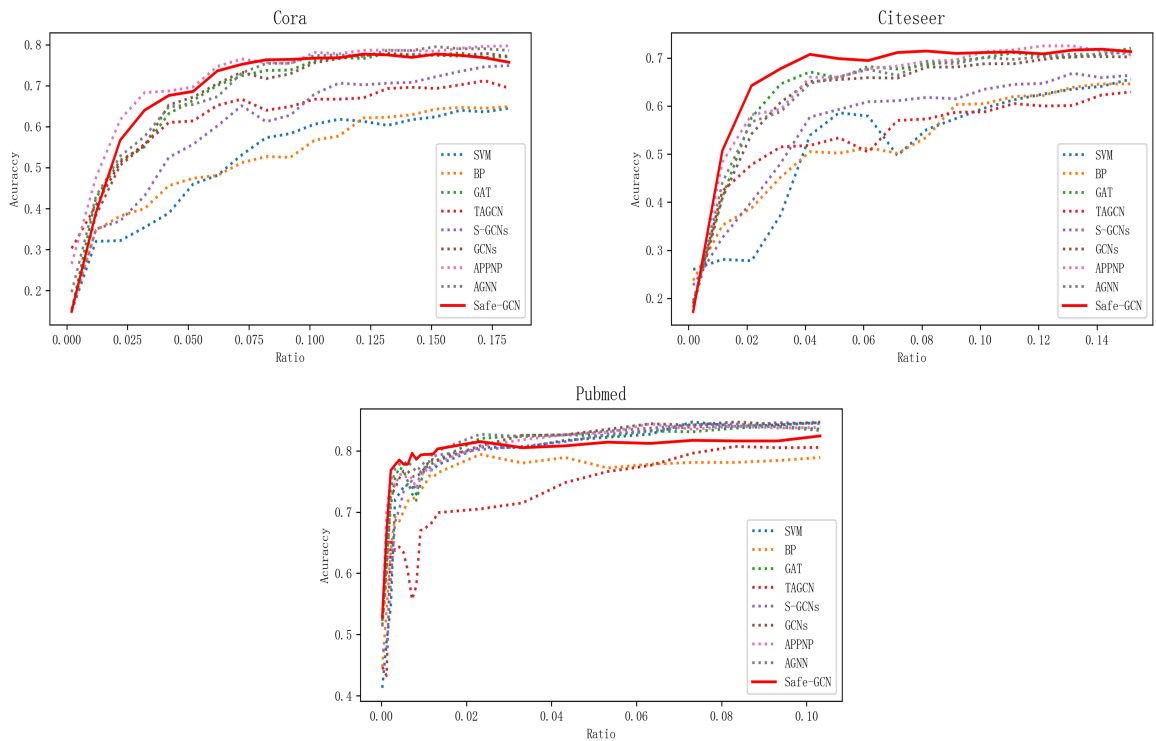}
\caption{Classification accuracy (\%) of the proposed method for different ratios of labeled data (a) Cora dataset. (b) Citeseer dataset. (c) Pubmed dataset}
\label{fig:2}
\end{figure}

Since the number of labeled data has an effect on the accuracy of the model, it is interesting to study the performance of the model with different numbers of labeled data. The number is increased or decreased from the original labeled dataset. In this section, we adjust the number of initially labeled data to study the performance of the proposed model from different dataset. The basic labeled dataset of Cora, Citeseer, and Pubmed are respectively 140, 120, 60, accounting for 0.2\%, 0.16\%, and 0.02\% of the total training data, respectively. In Fig \ref{fig:2}, the horizontal coordinate indicates the proportion of labeled data to the overall training data, and the vertical coordinate indicates the classification accuracy of the model.

\subsection*{The effect of Confidence thresholds}

\begin{figure}[ht]
\centering
\includegraphics[width=\linewidth]{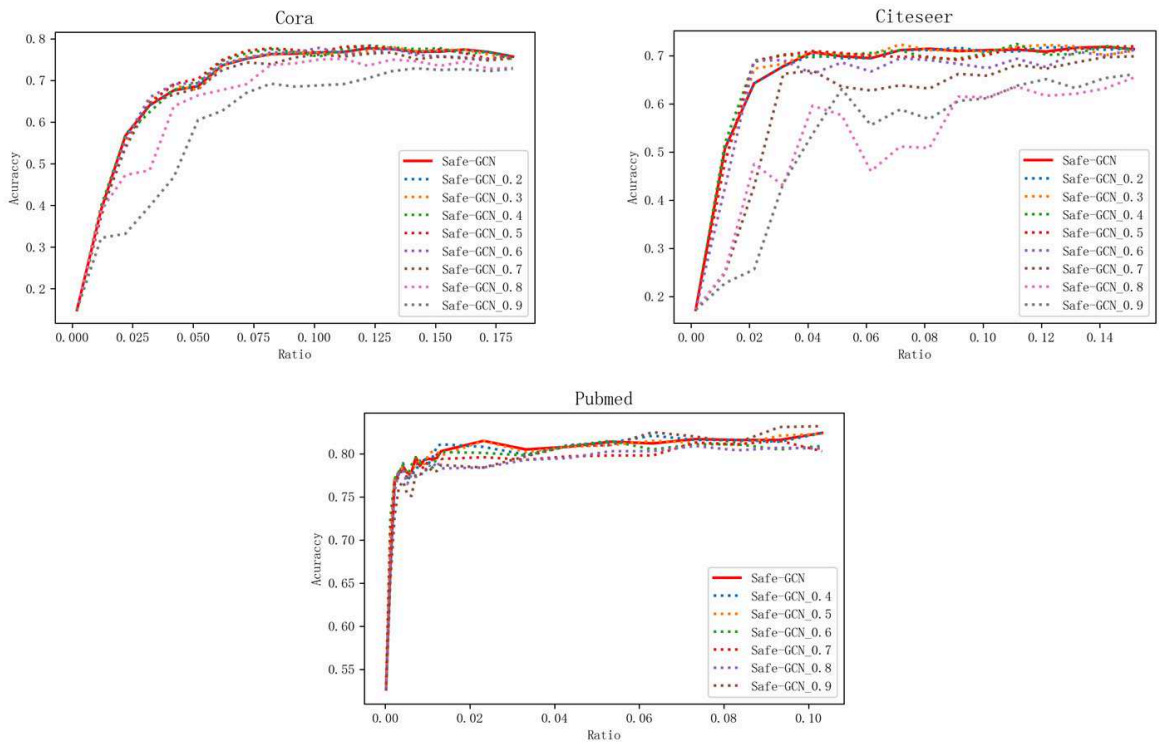}
\caption{Classification accuracy (\%) of the proposed method for different $\alpha$ of labeled data (a) Cora dataset. (b) Citeseer dataset. (c) Pubmed dataset.}
\label{fig:3}
\end{figure}

Hyperparameters are very important in machine learning, which can directly affect the performance of the model. In this paper, $\alpha$ as a hyperparameter denotes the confidence threshold used to determine the data security. Therefore, in order to study the effect of $\alpha$ on the model, we give a set of values to adjust it. The $\alpha$ of Cora, Citeseer were chosen from {[0.2,0.3,\ ...\ ,0.9]} and Pubmed were chosen from {[0.4,\ 0.5,\ ...\ ,0.9]}, respectively. The classification accuracy of the model for each of the three citation datasets at different $\alpha$ is illustrated in Fig \ref{fig:3}.

\subsection*{Experimental analysis}
From the results of all previous tables and pictures, we can conclude the following:

\begin{itemize}

\item[(1)]	From Table \ref{tab:2} and \ref{tab:3}, we can see that our model outperforms the other methods on all three citation datasets. In particular, it can improve more than 3\% on all three datasets compared to GCN. We also have different degrees of advancement compared to other methods. This indicates that high confidence data can enhance the predictive power of the model.

\item[(2)] The superiority of the proposed model in the case of small labeled data size is obvious as presented in Fig \ref{fig:2}, which indicates that the proposed model is applicable to the problem of few labels and can safely utilize a large number of unlabeled data to satisfy many real-world application scenarios. The advantages of the proposed model are not evident in the case of a large amount of labeled data, which is attributed to the fact that a large amount of labeled data can already describe the distribution of the data adequately.

\item[(3)] 	Fig \ref{fig:3} illustrates that in the Cora and Citeseer, the proposed model is insensitive in the ranges ${0.2,0.3,\ ...\ ,0.9}$ , ${0.4,\ 0.5,\ ...\ ,0.9}$  for parameter $\alpha$, respectively. In Pubmed, the proposed model is insensitive to the parameter $\alpha$ and even to the proportion of labeled data in the training data.

\end{itemize}

\section{Conclusion}
We propose a safe GCN framework. The model is based on the self-training framework, which utilizes embedding information of unlabeled data by learning S-GCN classifiers and GCN classifiers. Then obtains high-confidence unlabeled data using a confidence threshold-based data filtering condition, which is balanced to expand the labeled data and reduce the negative impact of risky unlabeled data. At the same time, the model combines supervised and semi-supervised information of data, which improves the security of unlabeled data than using only supervised or semi-supervised information. Therefore, our model can effectively reduce the risk of unlabeled data and safely use a large number of unlabeled data. In addition, our model is applicable to inductive learning, which extends the applicability of the model to some extent. \

In the future work, we will focus on the following directions: (1) more detailed risk classification of unlabeled data, and different risk levels of unlabeled data may have different effects on the model. (2) The quality of the model also affects the performance of the model, and methods to assess the quality of the model will be explored. (3) Reducing the time complexity of the model is of importance in realistic application scenarios.

\bibliographystyle{unsrt}
\bibliography{safe-gcn_cite}
\end{document}